\documentclass{llncs}
\usepackage{makeidx}  
\usepackage{graphicx}
\usepackage{subfigure}
\usepackage{amssymb,amsmath,bm}
\usepackage{mathrsfs}
\usepackage{booktabs}
\usepackage{amsmath}
\usepackage{mathtools}
\usepackage{subfig}
\usepackage[misc]{ifsym}
\usepackage{multirow}
\usepackage{bbding}
\usepackage{array}
\usepackage{float}
\usepackage{algorithm}
\usepackage{algorithmic}
\usepackage{lettrine}
\usepackage{cite}

\newcolumntype{C}[1]{>{\centering\let\newline\\\arraybackslash}m{#1}}

\usepackage{pbox}
\begin{document}
	\frontmatter          
	\title{Identification of primary angle-closure on AS-OCT images with Convolutional Neural Networks}
	
%
	
	\author{Chenglang Yuan$^{1,2}$ \and Cheng Bian$^{1}$ \and Hongjian Kang$^{1}$ \and Shu Liang$^{2}$ \and Kai Ma$^{1}$ \and Yefeng Zheng$^{1}$}
	\institute{Tencent YouTu Lab \\
		\email{tronbian@tencent.com} \and
		School of Biomedical Engineering, Health Science Center, Shenzhen University, Shenzhen, China \\
		\email{yuanchenglang@email.szu.edu.cn}}
	
	\maketitle
	
	\begin{abstract}
		Primary angle-closure disease (PACD) is a severe retinal disease, which might cause irreversible vision loss. In clinic, accurate identification of angle-closure and localization of the scleral spur’s position on anterior segment optical coherence tomography (AS-OCT) is essential for the diagnosis of PACD. However, manual delineation might confine in low accuracy and low efficiency. In this paper, we propose an efficient and accurate end-to-end architecture for angle-closure classification and scleral spur localization. Specifically, we utilize a revised ResNet152 as our backbone to improve the accuracy of the angle-closure identification. For scleral spur localization, we adopt EfficientNet as encoder because of its powerful feature extraction potential. By combining the skip-connect module and pyramid pooling module, the network is able to collect semantic cues in feature maps from multiple dimensions and scales. Afterward, we propose a novel keypoint registration loss to constrain the model's attention to the intensity and location of the scleral spur area. Several experiments are extensively conducted to evaluate our method on the angle-closure glaucoma evaluation (AGE) Challenge dataset. The results show that our proposed architecture ranks the first place of the classification task on the test dataset and achieves the average Euclidean distance error of 12.00 pixels in the scleral spur localization task.
	\end{abstract}
	
	\section{Introduction}
	
\begin{figure}[h]
		\centering
		\includegraphics[width=1\linewidth]{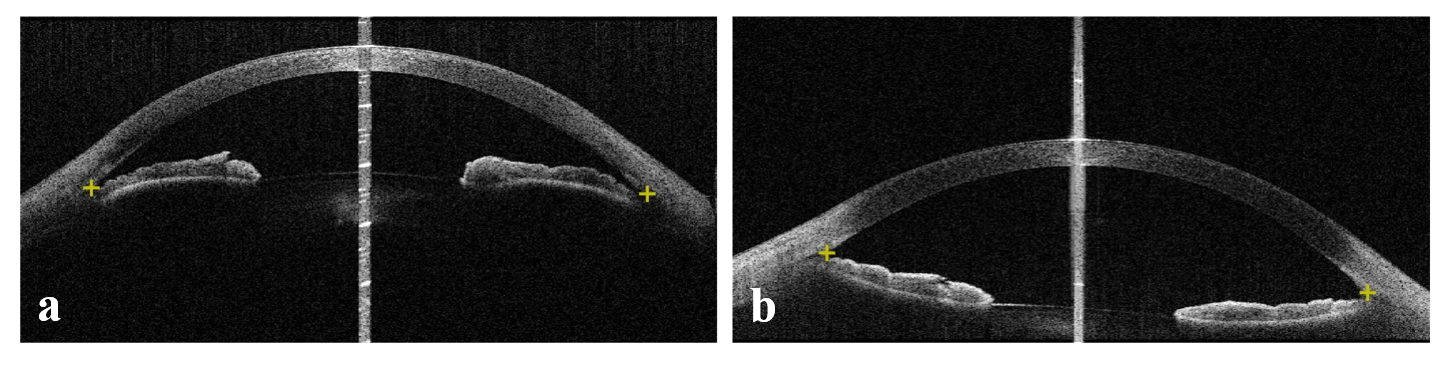}
		\caption{Two AS-OCT images with annotated scleral spur locations (the yellow marks): a) A closure angle OCT image. b) An open angle OCT image.}
		\label{fig:fig1}
		\vspace{-10.0mm}
	\end{figure}

	Primary angle-closure disease (PACD) is the primary type of glaucoma, which is caused by the abnormal anatomical structures of the chamber angle \cite{Nongpiur2011angle}. Up to year 2019, PACD comes to be the leading cause of blindness and affects a large number of people worldwide \cite{prum2016primary,quigley2006number}. Progression of PACD without diagnosis and appropriate treatment gradually leads to irreversible vision loss \cite{yu2019robust,guzman2013anterior}. Therefore, early diagnosis and intervention of PACD have a crucial influence on the proper management of the disease. In clinic, anterior segment optical coherence tomography (AS-OCT) is the frontier optical imaging technology to help ophthalmologists understand and identify the process of PACD. Concretely, identifying the status of anterior angles (open/closed) and localizing the position of scleral spurs (SS) on AS-OCT images form the conventional method to diagnose PACD. Fig.~\ref{fig:fig1} gives examples of angle-closure/normal AS-OCT images overlapped with SS annotations.
	
    However, this manual evaluation is time-consuming and subjective, which might lead to the unstable and inaccurate results due to experience of the operators. Meanwhile, the cumulative errors collected from speckle noise interference and significant changes in the appearance or intensity of target regions among different AS-OCT images, make it challenging to analyze the changes of anterior chamber anatomical structures. Particularly, compared to other tissue regions, SS is a tiny and ambiguous structure which makes it hard for ophthalmologists to locate the position precisely. Hence, it is essential to develop an automatic algorithm for accurate identification of the anterior angle and localization of the SS position.

   The classification of anterior angle based on AS-OCT images is an active research area. Previously, Nongpiur et al. \cite{Nongpiur2013angle} proposed a classification algorithm based on stepwise logistic regression that used a combination of six parameters obtained from a single horizontal AS-OCT scan to identify subjects with gonioscopic angle and achieved AUC of 0.96. Yan et al. \cite{Yan2013auto} proposed an image processing and machine learning based framework to localize and classify anterior chamber angle with OCT images. This framework achieved AUC of 0.92 and accuracy of 0.84. Although the performance of these above methods was within the acceptable error in the application of angle-closure identification, it is difficult for machine learning models to determine the category of anterior angles with poor image quality.

    Due to the capability to automatically learn a hierarchical representation of input data without much human involvement, the convolutional neural network has been widely applied to solve practical problems and achieves state-of-the-art performance in computer vision. In this paper, we propose a deep convolution network architecture utilizing OCT images to perform angle closure classification and SS localization with promising performance, thus providing a novel method for clinical diagnosis of PACD.
    \vspace{-6mm}
	\section{Methodology}
\begin{figure}[h]
		\centering
		\includegraphics[width=1\linewidth]{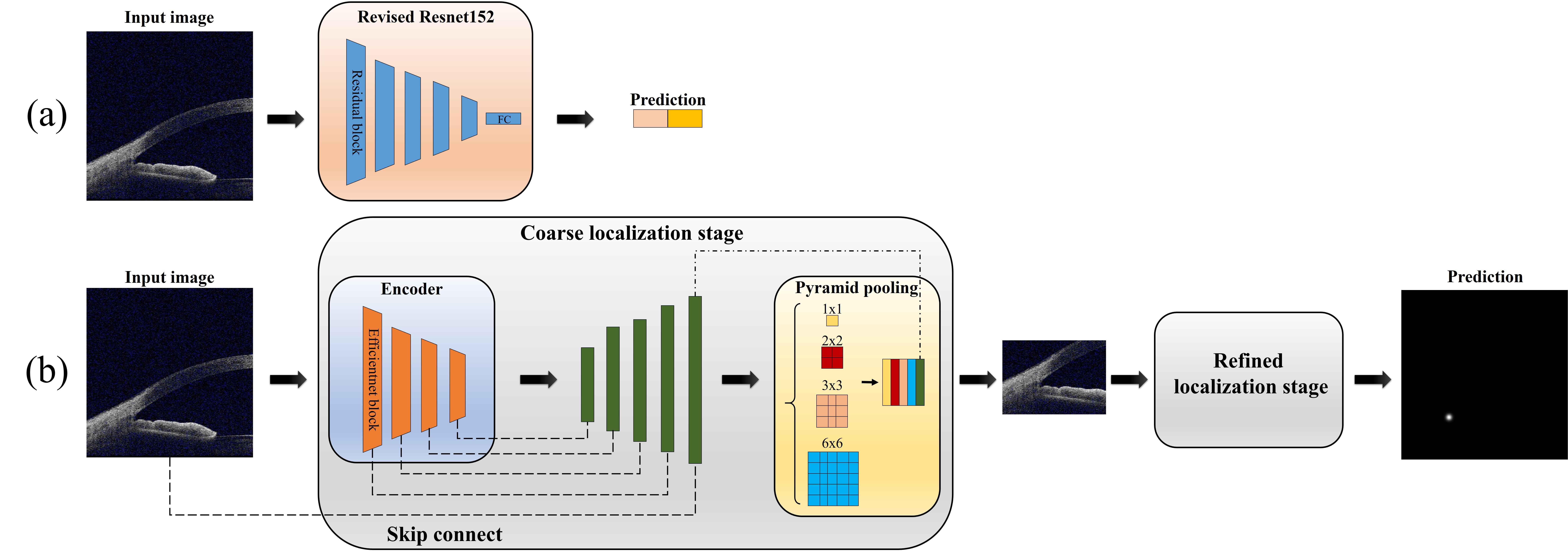}
		\caption{The proposed architecture. a) The angle closure classification network. b)The scleral spur localization network.}
		\label{fig:fig2}
		\vspace{-8mm}
	\end{figure}

	Fig.~\ref{fig:fig2} shows the proposed architecture. According to the rule of the AGE challenge \cite{petbfy1019}, our architecture includes an angle-closure classification task and an SS localization task. To avoid the mutual influence between two tasks, we develop the PACD classification model and the SS localization model, respectively. Since we assume that the key to the classification of closure angles and localization of SS is to make use of features extracted from surrounding regions of sclera, we split the full OCT images into left/right parts and only take one of them as input. This strategy reduces tremendous computational consumption and increases the amount of training data. Then, we design a deep residual network to identify the status of angles on OCT images (i.e., open or closure) in the angle-closure classification task.
	
	On the other hand, we adopt a two-stage progressively tuning strategy (the coarse and refined localization networks are identical) for the SS localization task. Concretely, we firstly utilize an efficient and effective encoder to learn and extract hierarchical features. After that, a skip-connect module and a pyramid pooling module (PPM) are implemented to capture the semantic features from multiple dimensionalities and scales. The semantic features are merged to infer the final response regions. Finally, the keypoint registration loss is introduced to guide the network focusing on the SS position.
	\vspace{-2mm}

	\subsection{Angle-closure classification network}
	Previous experiments showed that the depth of the network is essential for model performance \cite{lin2017focal}. However, when we stack the convolutional blocks excessively, the performance of model degrades rapidly after the network reaches a certain depth because of the vanishing gradient problem. To alleviate this issue, He et al.~\cite{he2016deep} proposed a residual learning framework to effectively promote the gradient update in backward propagation. For the same purpose, we employ ResNet152 as the backbone network architecture to perform the accurate identification of angle closure in our task. Referring to He et al.~\cite{he2019bag}, we develop two ResNet tweaks to improve the capacity of the angle-closure identification network. 
	
\begin{figure}[h]
		\centering
		\includegraphics[width=0.85\linewidth]{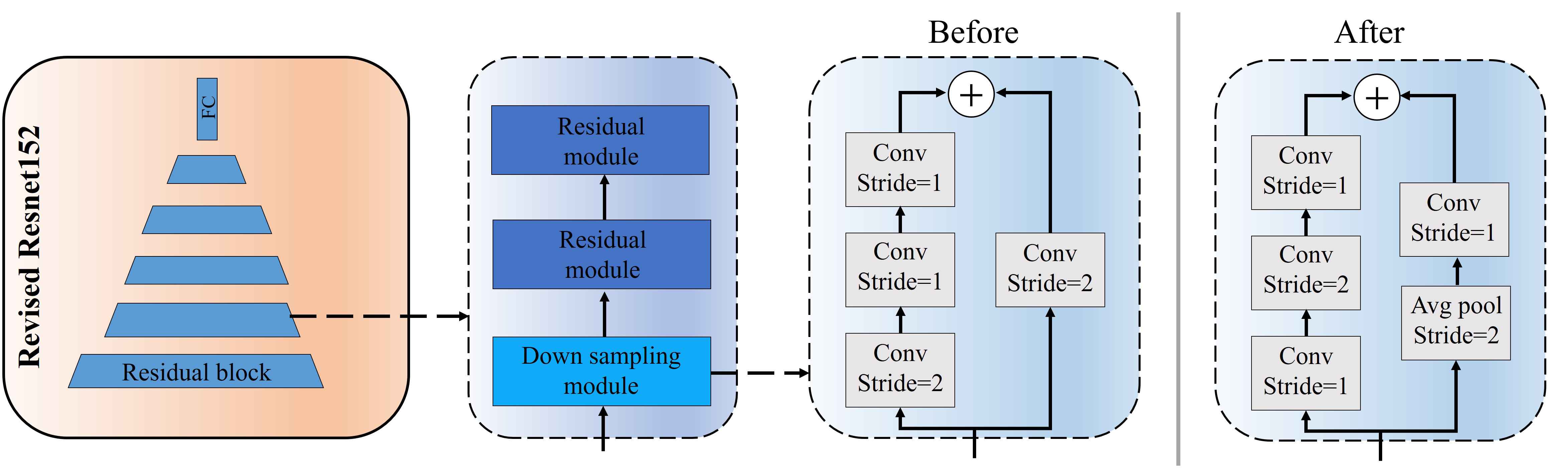}
		\caption{The details of ResNet tweaks.}
		\label{fig:fig3}
		\vspace{-8mm}

	\end{figure}

    The details of ResNet tweaks are shown in Fig. \ref{fig:fig3}. To alleviate the contextual information loss from the first convolution with a stride of 2 in down sampling module of Residual blocks, we switch the strides of the first two convolutions. Similarly, the residual connect mechanism in the down sampling module also ignores 3/4 of input feature maps. As shown in Fig. \ref{fig:fig3}, empirically, we insert a 3$\times$3 average pooling layer with a stride of 2 before the convolution layer. Moreover, we set the convolution stride to 1. All these tweaks work well in practice and promote identification performance as well as computational efficiency.
    \vspace{-4mm}
	\subsection{Scleral spur localization network }
	Instead of directly regressing the SS position, we utilize the classic keypoint detection technique introduced in \cite{zhou2019objects}. The ground-truth heatmap is a 2D Gaussian distribution centered at SS location. For the SS localization task, our network architecture contains an encoder, a skip-connect module and a pyramid pooling module.
	\subsubsection{Encoder.}
	Tan et al. \cite{tan2019efficientnet} identified that carefully balancing network depth, width, and resolution can lead to better performance. Based on this observation, they designed EfficientNet with a new scaling method that uniformly scales all dimensions of depth/width/resolution. Compared with ResNet, EfficientNet has excellent performance to catch the high-dimensional semantic information and achieves much better accuracy and efficiency. Therefore, we choose EfficientNet as our encoder. 
	\subsubsection{Skip-connect module and pyramid pooling module.}
    The SS is a tiny part, which is attached anteriorly to the trabecular meshwork and posteriorly to the sclera and the longitudinal portion of the ciliary muscle. Therefore, we need more contextual information, including sclera and other tissues around the sclera, to localize the SS. Inspired by U-Net \cite{ronneberger2015u} and PSPNet \cite{zhao2017pyramid}, we implement a skip-connect module and a PPM to enhance the utilization of multi-level and multi-scale information. Concretely, for the skip-connect module, we adopt four transposed convolutions with the stride of 2 to construct a short connection with the high-level and the corresponding low-level features from the encoder. In contrast, for the PPM, we set four pooling layers to obtain the multi-scale contextual features, then upsample these features to the original size and integrate it to predict the scleral spur’s location. It is worth noting that the bin size of these pooling layers is set to 1$\times$1, 2$\times$2, 3$\times$3 and 6$\times$6, respectively.
	\vspace{-3mm}
	\subsection{Loss function}
	After calculating the statistics of the original data, we find that the number of normal images is four times of the angle-closure images. In order to overcome the imbalance between normal and angle-closure samples, we employ a hybrid loss combining the focal loss \cite{lin2017focal} and F-beta loss. The focal loss is defined as:
	\begin{equation}
		\mathcal L_{focal} = - \alpha(1-y'_{t})^{\gamma}\log{y'_t}
	\end{equation}
	\begin{equation}
	y'_{t} = yy'+(1-y)(1-y')
	\end{equation}
	where  $y \in [0,1]$ specifies the ground-truth class; $y' \in [0,1]$ represents the predictions of the model; $\alpha$  is the hyper-parameter to balance the importance between two classes; Modulating factor $\gamma$ is utilized to force the model to mine the hard negative samples. The F-beta loss is defined as:
	\begin{equation}
	\mathcal L_{Beta} = 1-\frac{(1+\beta)\times \mathbf{TP}}{(1+\beta)\times\mathbf{TP}+\beta^2\times\mathbf{FN}+\mathbf{FP}}
	\end{equation}
	where $\beta$ is the regulatory factor to balance the weights; TP, FN and FP denote the number of true positives, false negatives, and false positives, respectively. Afterwards, we define the hybrid loss as:
	\begin{equation}
	\mathcal L_{hybrid} = \rho_{1}L_{focal}+\rho_{2}L_{Beta}
	\end{equation}
	where $\rho_1$ and $\rho_{2}$ are the weights for $L_{focal}$ and $L_{Beta}$, respectively.

	Considering the intensity and shape of SS regions in the localization task, we propose a keypoint registration (KR) loss to reduce error between prediction and ground truth. Since the SS location heatmap implicitly contains the intensity and shape information, our proposed KR loss can force the network to concentrate on the position of the heatmap and the intensity. Finally, our KR loss consists of the position and intensity regression parts, which can be defined as:
	\begin{equation}
	\mathcal L_{KR} = \rho_{3}(\frac{1}{N}\sum (y-y')^2)+\rho_{4}(1-2\sum\frac{ |yy'|}{|y|+|y'|})
	\end{equation}
	where $y'$ and $y \in [0,1]$ represent the model estimated probability and ground truth, respectively; N represents the number of pixels; $\rho_{3}$ and $\rho_{4}$ are the weights to regulate two losses.

	\section{Experiments} 
	\subsubsection{Experiment materials:} We evaluate our method on the AGE challenge dataset of MICCAI 2019\cite{petbfy1019}. There are 1600 OCT images of size 998$\times$2130 pixels provided for quantitative analysis. We split the original image into left and the right parts, then resize the split image to 256$\times$256 pixels for the angle-closure classification task. In contrast, for the SS localization task, we resize the split image to 499$\times$499 pixels as input for the first stage. For the second localization stage, we crop the coarse region of 384$\times$288 as input from the previous stage. We split the whole dataset with ratio of 80/20 for training/testing. 
	
	\subsubsection{Implementation details:} We implemented our architecture in PyTorch using an NVIDIA Tesla P40 GPU. The Adam optimizer and cosine learning rate decay strategy are applied to update the model weights (the batch size in localization task is 27 and the batch size in classification task is 72, both initial learning rates are 0.001). Using the grid search strategy, $\alpha$, $\gamma$, $\beta$, $\rho_{1}$, $\rho_{2}$, $\rho_{3}$, $\rho_{4}$ are set to 2, 2, 2, 0.5, 0.5, 100, 0.2, respectively. Several state-of-the-art networks are implemented for extensive comparisons.
	\vspace{-6mm}
	\begin{table}[h]
		\centering
		\caption{Comparisons between the proposed method and other methods in angle closure classification.}\label{table:t1}
		\vspace{2mm}
		\begin{tabular}{p{2cm}<{\centering}|p{1cm}<{\centering}|p{1cm}<{\centering}|p{1cm}<{\centering}|p{1cm}<{\centering}}
			\toprule[2pt]
			\multicolumn{1}{c|}{\bf{Method}}&\multicolumn{1}{c|}{\bf{AUC}}&\multicolumn{1}{c|}{\bf{SEN}}&\multicolumn{1}{c|}{\bf{SPE}}&\multicolumn{1}{c}{\bf{ACC}}\\
			\hline
			ResNet152 & 0.99 &  1.00 & 0.99 & 0.99 \\
			\textbf{Proposed} & \textbf{1.00} &  \textbf{1.00} & \textbf{1.00} &\textbf{1.00} \\
			\bottomrule[2pt]
		\end{tabular}
	\end{table}
	\vspace{-16mm}

	\begin{table}[htp]
	\centering
	\caption{Comparisons between the proposed method and other methods in scleral spur localization.}\label{table:t2}
	\begin{tabular}{p{2cm}<{\centering}|p{2.5cm}<{\centering}|p{2.5cm}<{\centering}|p{2.5cm}<{\centering}}
		\toprule[2pt]
		\multicolumn{1}{c|}{\bf{Method}}&\multicolumn{1}{c|}{\bf{\pbox{2.5cm}{Left ED error\\ (pixel)\centering}}}&\multicolumn{1}{c|}{\bf{\pbox{2.5cm}{Right ED error\\ (pixel)\centering}}}&\multicolumn{1}{c}{\bf{\pbox{2.5cm}{Avg ED error\\ (pixel)\centering}}}\\
		\hline
		U-Net \cite{ronneberger2015u}& 12.57 &  15.30 & 13.93  \\
		ResUNet & 12.14 &  14.17 & 13.15  \\
		UNet++\cite{Zhou2018Unet++} & 10.60 &  14.39 & 12.50  \\
		\textbf{Proposed} & \textbf{10.28} & \textbf{13.73} & \textbf{12.00}  \\
		\bottomrule[2pt]
	\end{tabular}
	\end{table}
	\vspace{-10mm}

	\begin{table}[!h]
		\centering
		\caption{\small Ablation study of the proposed method in scleral spur localization.}\label{table:t3}
		\vspace{2mm}
		\begin{tabular}{p{1.5cm}<{\centering}|p{1.5cm}<{\centering}|p{1.5cm}<{\centering}|p{2.5cm}<{\centering}|p{2.5cm}<{\centering}|p{2.5cm}<{\centering}}			
			\toprule[2pt]
			\multicolumn{1}{C{1.5cm}|}{\bf{Encoder}} & \multicolumn{1}{C{1.5cm}|}{\bf{PPM}} & 	\multicolumn{1}{C{1.5cm}|}{\bf{KR loss}} & \multicolumn{1}{C{2.5cm}|}{\bf{\pbox{2.5cm}{Left ED error\\ (pixel)\centering}}} &	\multicolumn{1}{C{2.5cm}|}{\bf{\pbox{2.5cm}{Right ED error\\ (pixel)\centering}}} & \multicolumn{1}{C{2.5cm}}{\bf{\pbox{2.5cm}{Avg ED error\\ (pixel)\centering}}}\\
			\hline
			& & & 11.79 & 14.99 & 13.39  \\
			\checkmark & & & 10.42&14.67&12.54  \\
			\checkmark & \checkmark & & 10.40&14.02&12.21  \\
			\checkmark & \checkmark & \checkmark & 10.28&13.73&12.00  \\
			\bottomrule[2pt]
		\multicolumn{6}{l}{$^{*}$\footnotesize{The default encoder contains four convolution and pooling layers.}} \\
	\end{tabular}
	\vspace{-6mm}
	\end{table}
		
%
%
    
	\subsubsection{Quantitative and Qualitative Analysis:} 
\begin{figure}[h]
		\centering
		\includegraphics[width=1.0\linewidth]{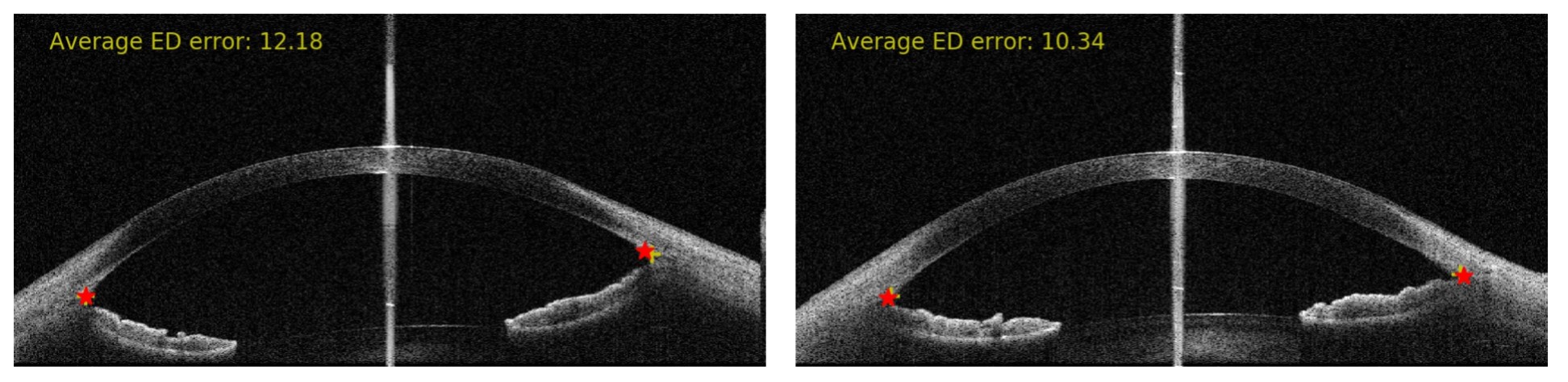}
		\caption{Some visualized sample results. The yellow marks represent the groundtruth, while the red marks represent the prediction outputs.}
		\label{fig:fig4}
		\vspace{-8mm}
	\end{figure}

We employ AUC, sensitivity (SEN), specificity (SPE) and accuracy (ACC) to evaluate the proposed method in the angle-closure classification task. The left/right Euclidean distance (ED) error and average Euclidean distance error are used in the SS localization task. As shown in Table \ref{table:t1} and Table \ref{table:t2}, we can observe that, the revised ResNet152 makes use of more context information, thus exhibiting higher performance than original ResNet152 in all metrics. For SS localization, due to the explicit exploration of muli-dimension and multi-scale semantic features, the proposed method obtains better results than other networks. We further conduct an ablation study on the encoder, PPM and KR loss in Table \ref{table:t3}. There occurs a significant improvement (about 0.85 average ED error) when we select EfficientNet as the encoder. This demonstrates the importance of extracting effective semantic features. Moreover, PPM brings another 0.33 improvement in the average ED error because of the collected semantic cues from the multi-scale feature maps. The KR loss which guides the network to determine the SS position, achieves state-of-the-art performance. Some sample results are visualized in Fig. \ref{fig:fig4}.
	
	\vspace{-2mm}

	\vspace{-3mm}
	\section{Conclusions}
	In this study, we proposed a new architecture to automatically and accurately identify open/closed angles and determine the position of SS. Both ResNet and EfficientNet can provide a large amount of semantic information to mine the intrinsic relationship between images and clinical characteristics. The utilization of skip-connect module and PPM improves the model performance through combining the global and local properties from different dimensions and sizes to recognize target regions. Two specific loss functions boost training efficacy and push the network for further improvement. Verified on the final test dataset, our method presents promising performance and ranks the third place in the overall leadboard. 
	
	
	%
	\bibliographystyle{splncs}
	\bibliography{refs}	
	
\end{document}